\documentclass[conference]{IEEEtran}
\IEEEoverridecommandlockouts
\usepackage{cite}
\usepackage{amsmath,amssymb,amsfonts}
\usepackage{algorithmic}
\usepackage{graphicx}
\usepackage{textcomp}
\usepackage{xcolor}
\usepackage{flushend}

\def\BibTeX{{\rm B\kern-.05em{\sc i\kern-.025em b}\kern-.08em
    T\kern-.1667em\lower.7ex\hbox{E}\kern-.125emX}}
\begin{document}

\title{An Optimized Toolbox for Advanced Image Processing with Tsetlin Machine Composites\\
}

\author{\IEEEauthorblockN{Ylva Grønningsæter}
\IEEEauthorblockA{\textit{Centre for AI Research} \\
\textit{University of Agder}\\
Grimstad, Norway \\
0009-0001-7855-069X}
\and
\IEEEauthorblockN{Halvor S. Smørvik}
\IEEEauthorblockA{\textit{Centre for AI Research} \\
\textit{University of Agder}\\
Grimstad, Norway  \\
0009-0007-1961-4466}
\and
\IEEEauthorblockN{Ole-Christoffer Granmo}
\IEEEauthorblockA{\textit{Centre for AI Research} \\
\textit{University of Agder}\\
Grimstad, Norway \\
0000-0002-7287-030X}
}

\maketitle

\begin{abstract}
The Tsetlin Machine (TM) has achieved competitive results on several image classification benchmarks, including MNIST, K-MNIST, F-MNIST, and CIFAR-2. However, color image classification is arguably still in its infancy for TMs, with CIFAR-10 being a focal point for tracking progress. Over the past few years, TM's CIFAR-10 accuracy has increased from around 61\% in 2020 to 75.1\% in 2023 with the introduction of Drop Clause. In this paper, we leverage the recently proposed TM Composites architecture and introduce a range of TM Specialists that use various image processing techniques. In addition, we conduct a rigorous hyperparameter search, where we uncover optimal hyperparameters for several of the TM Specialists. The result is a toolbox that provides new state-of-the-art results on CIFAR-10 for TMs with an accuracy of 82.8\%. In conclusion, our toolbox of TM Specialists forms a foundation for new TM applications and a landmark for further research on TM Composites in image analysis.

\end{abstract}

\begin{IEEEkeywords}
Machine Learning, Tsetlin Machine, Image Classification, Image Processing Techniques
\end{IEEEkeywords}

\section{Introduction}
\label{sec:introduction}
The Tsetlin Machine (TM) \cite{granmo_tsetlin_2021} is a promising machine learning algorithm that has attracted increasing attention since its introduction in 2018. Recent research addresses natural language processing \cite{yadav_human-level_2021, bhattarai_measuring_2021, saha_using_2023}, intrusion detection \cite{gunvaldsen_towards_2023}, energy-efficient hardware design for internet of things \cite{wheeldon_learning_2020},
and image classification \cite{sharma_drop_2023, maheshwari_redress_2023, zhang_interpretable_2023}. The TM provides energy efficiency and interpretability, primarily due to its utilization of propositional logic \cite{granmo_tsetlin_2021}. As such, TMs address the black-box nature and complexity of deep neural networks, with their lack of interpretability and significant computational cost \cite{saleem_explaining_2022, sze_efficient_2017}. Furthermore, the TM has been able to achieve competitive performance on several benchmarks, such as optimal move prediction in board games \cite{granmo_tsetlin_2021}, text classification \cite{berge_using_2019}, and image classification \cite{granmo_convolutional_2019, maheshwari_redress_2023}. In terms of image classification, the TM has achieved state-of-the-art performance on the image datasets MNIST, K-MNIST, F-MNIST, and CIFAR-2 \cite{granmo_convolutional_2019, maheshwari_redress_2023}. However, color image classification is arguably still in its infancy for TMs, with CIFAR-10 being a focal point for tracking progress. During the past few years,~TM's~CIFAR-10 accuracy has increased from around 61\% in 2020 \cite{mathisen_analysis_2020} to~75.1\% in~2023 with the introduction of Drop Clause \cite{sharma_drop_2023}. Drop Clause is a technique for TMs that randomly drop clauses to avoid overfitting when learning patterns. The result is reduced training time as the TM has fewer clauses to check if the pattern matches.
Recently, the \textit{TM Composites} architecture introduced a plug-and-play approach to combining different TMs into a team~\cite{granmo_tmcomposites_2023}, meaning that multiple TMs can be independently trained and added to the TM Composite. By combining four different TMs, the paper demonstrated how multiple TMs can enhance each other's performance with a simple interaction scheme, attaining competitive performance with Drop Clause~\cite{granmo_tmcomposites_2023}. As such, TM Composites offers a promising strategy for boosting the accuracy of TMs. The lack of high-accuracy interpretable techniques in image analysis is problematic, in particular for high-stakes applications, including medical ones \cite{rudin_stop_2019}. Compared to the Vision Transformer proposed by Dosovitskiy et al. \cite{dosovitskiy_image_2021}, which has 632 million parameters and achieves an accuracy of 99.9\%, current interpretable approaches fall short. However, with 632 million parameters the Vision Transformer is incomprehensible to humans. Furthermore, the estimated carbon emissions produced by training a similar Transformer model with 213 million parameters have been compared with the lifetime carbon emissions of five cars \cite{strubell_energy_2019}. There is a need for further research to reduce the accuracy gap between interpretable and energy-efficient techniques and state-of-the-art deep learning. In light of these considerations, this paper aims to investigate how the TM's capabilities in image classification can be further enhanced by building upon the TM Composites architecture. Our paper contributions are as follows:
\begin{itemize}
     \item We introduce an approach to image preprocessing and processing for TM Specialists, focusing on enhancing their capability to handle color images. This includes the implementation of various image processing techniques.
    \item We conduct an extensive hyperparameter search for the TM Specialists, where we discover optimal hyperparameters for several of the TM Specialists.
    \item  We analyze the scalability of our TM Composites on CIFAR-10, investigating how the number of clauses per TM Specialist affects accuracy.
    \item We present a toolbox of 22 TM Specialists trained on augmented and non augmented data utilizing seven image Booleanization techniques that results in new state-of-the-art results for TMs on CIFAR-10 with an accuracy of~82.8\%.
    
\end{itemize}

The remainder of the paper is organized in the following manner. In Section~\ref{sec:tsetlin_machine}, we explain the fundamentals of the TM. In addition, we provide the basics of the TM Composites architecture. In Section~\ref{sec:implementation}, we introduce our image processing toolbox, covering each of the 22 TM Specialists in detail, including the underlying image processing approaches. In Section~\ref{sec:results}, we provide the basis for the experiments and present our results on CIFAR-10. 
Then, in Section~\ref{sec:related-work}, we highlight previous research done on TMs concerning image classification in color images. Finally, in Section~\ref{sec:conclusion}, we conclude our study and propose opportunities for further work.

\section{Tsetlin Machine Fundamentals}
\label{sec:tsetlin_machine}
\subsection{Tsetlin Automaton}
\label{subsec:tsetlin_automaton}

\begin{figure}[ht]
    \centering
    \includegraphics[width=\linewidth]{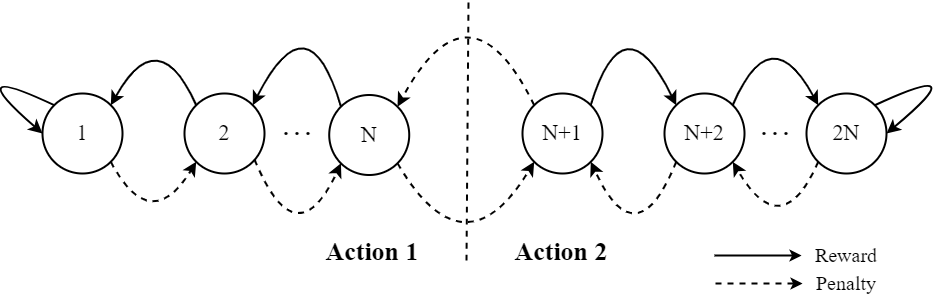}
    \caption{State transition diagram for a two-action Tsetlin Automaton with $2N$ states \cite{granmo_tsetlin_2021}.}
    \label{fig:learning_automaton}
\end{figure}

The \textit{Tsetlin Automaton} (TA) is a type of learning automaton designed to identify and execute the most efficient action in an unknown and stochastic environment \cite{granmo_tsetlin_2021}. To achieve this, a two-action TA performs its actions one-by-one, observing the effect each action has on the environment. That is, every action receives a reward or a penalty, which consequently alters the state of the TA. The TA is not aware of the reward probabilities for the actions, and thus has to learn through trial-and-error. Additionally, the reward probability for a specific action does not have to be constant and could potentially change. Fig.~\ref{fig:learning_automaton} shows a two-action TA, its states, and the state transitions achieving learning.

This TA has $2N$ states, where states $1$ to $N$ initiate Action 1, and states $N+1$ to $2N$ initiate Action 2 \cite{granmo_tsetlin_2021}. The rewarding of an action transitions the TA state in the direction of that action. That is, the more certain a TA is of an action, it increasingly transitions toward either state $1$ for Action 1 or state $2N$ for Action 2. However, if the TA receives a penalty, it transitions towards the middle states $N$ and $N+1$. This is how the TA is able to learn the most efficient action to take when working in an unknown and stochastic environment. 

\subsection{Tsetlin Machine}
\label{subsec:tsetlin_machine}

The TM comprises numerous TAs, where each TA is responsible for the inclusion or exclusion of a specific literal~$l_k$ in a clause \cite{granmo_tsetlin_2021}. In this context, inclusion corresponds to Action 1, while exclusion corresponds to Action 2. Using the Boolean input feature vector $X = [x_1,x_2,...,x_o]$, the TM predicts the class of the input, selected among $m$ different classes \cite{granmo_tmcomposites_2023}. Using the features in $X$, the literal set, $L = \{x_1,x_2,...,x_o,\neg x_1, \neg x_2, \ldots,\neg x_o\}$, is constructed by combining the original input feature vector and the negation of the input feature vector.

In addition, the TM utilizes Boolean AND operations on literals to build sub-patterns. These sub-patterns are commonly referred to as \textit{conjunctive clauses}. A TM utilizes a group of~$n$ conjunctive clauses for each class. Here, $n$ is a hyperparameter set by the user.
Half of the conjunctive clauses are assigned positive polarity $(+)$, while the remaining half receive negative polarity $(-)$.
Formally, each clause $C_j^{i,p}$, where $i \in \{1,2,...,m\}$ is the index of the class, $p\in\{+,-\}$ is the polarity of the clause, and $j\in\{1,2,...,n/2\}$ is the index of the clause, can be expressed as:

\begin{equation}
\label{eq:clause}
   C_j^{i,p}(x) = \bigwedge\limits_{l_k \in L_j^{i,p}}l_k.
\end{equation}
In \eqref{eq:clause}, $L_j^{i,p}$ is a subset of the literal set $L$ that represents what literals $l_k$ the clause $C_j^{i,p}(x)$ should include \cite{granmo_tmcomposites_2023}.

A prediction is then made by applying all clauses for each class on the input feature vector. If a clause with positive polarity evaluates to 1 for the class, then the clause adds one vote to the class sum for the class. In contrast, if a clause with negative polarity evaluates to 1 for the class, then the clause subtracts one vote from the class sum for the class. By adding the positive votes together and subtracting the negative votes, the class sum for the class in question is created. Subsequently, the class is predicted by identifying the largest class sum among the individual class sums:
\begin{equation}
\label{eq:classification}
    \hat{y} = argmax_i \biggl( \sum_{j=1}^{n/2} C_j^{i,+}(x) - \sum_{j=1}^{n/2} C_j^{i,-}(x) \biggr).
\end{equation}
For further insights into how TMs learn clauses from data, please see the original TM paper \cite{granmo_tsetlin_2021}. The focus of the current paper is exclusively on the classification process.

\subsection{Tsetlin Machine Composites}
\label{subsec:tmcomposite}

The \textit{TM Composites} is an architecture that combines multiple TMs $t \in \{1, 2,...,r\}$ and uses their combined learned knowledge to predict the class by combining the class sums~$c^i_{t,f}$ of the different TMs $t$ \cite{granmo_tmcomposites_2023}. That is,
the TM Composites architecture is plug-and-play, meaning that TMs $t$ can be independently trained and then added to a TM Composite at any moment and in any combination. 
This is achieved by defining the input $X_f \in \mathcal{X} = \{X_1, X_2, \ldots,X_z\}$ and extending the calculation of the class sums from \eqref{eq:classification}. By doing this, the class sums of multiple TMs $t$ can be processed: 

\begin{equation}
\label{eq:composite_classification}
    c^i_{t,f} = \sum_{j=1}^{n/2} C_{t,j}^{i,+}(x_f) - \sum_{j=1}^{n/2} C_{t,j}^{i,-}(x_f).
\end{equation}

In \eqref{eq:composite_classification}, $i$ is the class index, $t$ is the TM index, $f$ is the input index for $x_f$, and $x_f$ is a single input from the input space~$X$~\cite{granmo_tmcomposites_2023}. A normalization constant is calculated for each class sum:

\begin{equation}
\label{eq:calculating_class_sums}
    \alpha_t = max_{f,i}(c^i_{t,f})-min_{f,i}(c^i_{t,f}).
\end{equation}

Finally, the class sums are normalized with the constant from \eqref{eq:calculating_class_sums}, so that each input $x_f$ gets a composition of normalized class sums, by adding them together \cite{granmo_tmcomposites_2023}. To obtain the predicted class, argmax is used on this composition of class sums:

\begin{equation}  
\label{eq:sum_class_sums}
    \hat{y}_f = argmax_i \Biggl( \sum_{t=1}^{r} \frac{1}{\alpha_t} c_{t,f}^{i} \Biggr).
\end{equation}
 
\section{A Toolbox for Tsetlin Machine Image Processsing}
\label{sec:implementation}
\begin{figure}[ht]
    \centering
        \includegraphics[width=\linewidth]{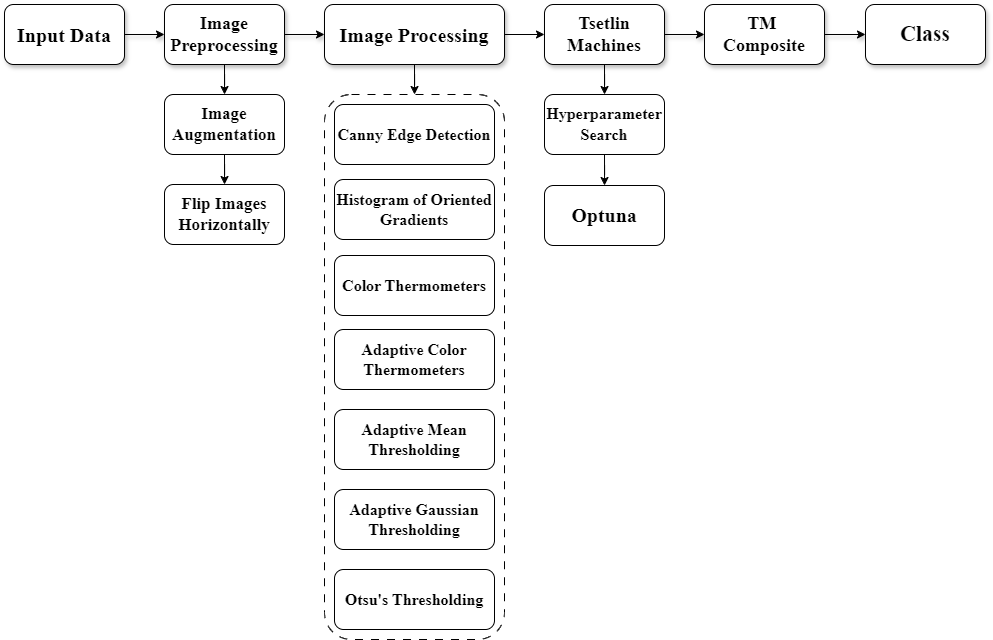}
        \caption{Proposed workflow for image classification with Tsetlin Machine Specialists utilizing image processing techniques.}
    \label{fig:method}
\end{figure}

We propose an approach for classifying images with TM Composites that consists of the four steps and substeps outlined in Fig.~\ref{fig:method}.
The process starts by augmenting the images in the input data. This involves duplicating each image in the dataset and then applying a horizontal flip to the duplicates, thereby enhancing the diversity and size of the dataset.

As covered in Section \ref{sec:tsetlin_machine}, the TM operates on Boolean input, and the modular nature of the TM Composites architecture facilitates the combination of multiple TMs. This setup allows for the application of diverse image processing techniques tailored to each TM within the composite. Our main goal is to create a toolbox of  image processing techniques, tailored for TM-based learning. Each technique provides complementary views on the data, selected to form a team of diverse TM Specialists, as proposed in~\cite{granmo_tmcomposites_2023}. In doing so, each TM can effectively contribute its unique strengths to the collective performance of the TM Composite it is part of. Given the significant variation in input images, it may be advantageous to have a team of specialists adept at optimally Booleanizing each specific type of input. 

Our toolbox consists of seven complementary image Booleanization techniques for the TM Specialists. These techniques are founded on Canny edge detection, adaptive mean thresholding, adaptive Gaussian thresholding, Otsu's thresholding, Histogram of Oriented Gradients (HOG), color thermometers, and adaptive color thermometers. These techniques were selected because they have demonstrated success in previous TM research \cite{mathisen_analysis_2020, granmo_tmcomposites_2023}, except for adaptive color thermometers, which is a novel addition. For further information, please refer to Section \ref{sec:related-work}.

After the images are Booleanized using one of the aforementioned techniques, the TM Specialists undergo an extensive hyperparameter search with Optuna\footnote{Optuna is an open-source hyperparameter optimization framework that automates the process of finding effective hyperparameters for machine learning models \cite{akiba_optuna_2019}.}. The hyperparameter search is essential for discovering optimal hyperparameters for each TM. Subsequently, the TM Specialists are retrained with optimal hyperparameters and added to a TM Composite. A TM Specialist is created for each image processing technique and each dataset.

The Booleanization and image processing techniques are explained in more detail in the following sections.

\subsection{Canny Edge Detection}
The first image processing technique is Canny edge detection. Canny edge detection is an algorithm that comprises five steps: applying a Gaussian filter, gradient calculation, applying non-maximum suppression, double thresholding, and performing edge tracking by hysteresis \cite{canny_computational_1986}.

To accommodate the multi-channel nature of color images, we adapted this algorithm to separately process the RGB channels of each image in our dataset. In other words, we process each channel independently as a separate grayscale image. This process begins with applying a Gaussian filter to smooth the input image, reducing noise and potential false edges. Following this, the gradient magnitude and direction for each pixel are calculated, employing Sobel operators to approximate the derivatives. The gradients, \(G_x\) and \(G_y\), are derived through the process of convolving the smoothed grayscale image, denoted as \(I_c\), where \(c \in \{R,G,B\}\), with a pair of \(3\times3\) convolution kernels:

\begin{equation}
\label{eq:gradient-x}
G_x = \begin{bmatrix}
-1 & 0 & 1\\
-2 & 0 & 2\\
-1 & 0 & 1\
\end{bmatrix} * I_c,
\end{equation}

\begin{equation} 
\label{eq:gradient-y}
G_y = \begin{bmatrix}
1 & 2 & 1\\
0 & 0 & 0\\
-1 & -2 & -1\
\end{bmatrix} * I_c.
\end{equation}

Above, $G_x$ and $G_y$ correspond to the gradients in the horizontal and vertical directions, respectively. The gradient magnitude, $G_c$, and orientation, $\theta_c$, at each pixel are then calculated as follows:

\begin{equation}
\label{eq:magnitude}
G_c = \sqrt{G^2_x + G^2_y},
\end{equation}

\begin{equation}
\label{eq:orientation}
\theta_c = \tan^{-1}\left(\frac{G_y}{G_x}\right).
\end{equation}

The subsequent stage involves non-maximum suppression, which thins the edges by retaining only the pixels that represent the local maxima of the gradient magnitude in the orientation of the gradient.

Finally, the algorithm employs double thresholding to differentiate between strong and weak edges, followed by edge tracking by hysteresis. Here, pixels with gradient magnitudes exceeding the high threshold are marked as strong edges, and those under the low threshold are suppressed. Pixels with gradients between these thresholds are classified as weak edges and are only retained if they are connected to strong edges, ensuring the detection of relevant structures while eliminating noise.

The output is then Booleanized, with edge pixels set to 1 and non-edge pixels set to 0, creating a Boolean edge map of the original images.

\subsection{Histogram of Oriented Gradients}

Histogram of Oriented Gradients (HOG) is a feature descriptor algorithm that converts an image of $height \times width \times channels$ to a feature vector of length $n$ \cite{dalal_histograms_2005}. The first step of HOG is calculating the gradients $G_x$ and $G_y$ for each pixel in an image $I$:
\begin{equation}
    G_x = \begin{bmatrix}
        -1 & 0 & 1
    \end{bmatrix} * I,
\end{equation}

\begin{equation}
    G_y = \begin{bmatrix}
        -1 \\ 
        0\\ 
        1\
    \end{bmatrix} * I.
\end{equation}

Subsequently, the magnitude and orientation of the gradients $G_x$ and $G_y$ are calculated using \eqref{eq:magnitude} and \eqref{eq:orientation}. The image is then divided into $n\times n$ cells, where a HOG is calculated for each cell. 
After the HOG is calculated for each cell, the $n\times n$ cells are normalized with a $m\times m$ block. Then the $m\times m$ blocks are concatenated to a HOG feature vector. Following the extraction of HOG features, a Booleanization process is applied to further refine the feature set for classification. This process involved the conversion of the HOG feature vectors into Boolean format, where each feature value was thresholded against a specific criterion. For each image, its feature descriptor is computed and each element of the descriptor is thresholded. If the element's value is greater than or equal to 0.1, it is set to 1 otherwise, it is set to 0.

\subsection{Adaptive Gaussian Thresholding}

The adaptive Gaussian thresholding technique begins by selecting a block size for each pixel that defines the local region around it \cite{ghaziasgar_computer_2020}. For each pixel, a threshold is calculated based on a Gaussian-weighted mean of the pixel values in the block. This method allows the threshold to dynamically adapt to the local luminance of the image; thus, accommodating different lighting conditions within the same image. As with Canny, each channel of the color image is processed independently as a separate grayscale image. The threshold $T$ for a given pixel $x,y$ is computed as follows:

\begin{equation}
T(x, y) = \mathit{WM}(x,y) - C.
\end{equation}

Here, $\mathit{WM}$ is the Gaussian-weighted mean of the pixel intensities in the block of $(x,y)$, and $C$ is a constant subtracted to fine-tune the thresholding sensitivity. 
Following the calculation of these adaptive thresholds, each pixel is compared against its local threshold. Pixels with intensity values greater than their corresponding thresholds are set to 1, whereas those with lower values are set to 0. 

\subsection{Adaptive Mean Thresholding}

Adaptive Mean Thresholding calculates the threshold for each pixel based on the mean intensity of pixels within a specified neighborhood around it \cite{gonzalez_digital_2007}, as opposed to adaptive Gaussian thresholding which utilizes a Gaussian-weighted mean to account for the proximity of neighboring pixels. Similar to the aforementioned methods, each channel is processed independently as a separate grayscale image. The threshold $T$ for each pixel $x,y$ is determined as follows for each channel~$c$:

\begin{equation}
T(x, y) = \mu(x, y) - C.
\end{equation}

Above, $\mu(x,y)$ represents the mean intensity value of the pixels within the pixel block, and $C$ is a constant that is subtracted to control the sensitivity of the thresholding process. 
Here, each pixel is compared against its adaptive threshold: pixels with intensity values greater than their corresponding thresholds are set to 1, while those with lower values are set to 0.

\subsection{Otsu's Thresholding}

Otsu's thresholding is a method that finds the optimal threshold intensity value by maximizing the between-class variance $\sigma^2_B$ in a given grayscale image \cite{otsu_threshold_1979}. Similar to the aforementioned image processing techniques, each channel of the image is processed independently as a separate grayscale image. In an image, each pixel is represented by $L$ gray levels~$[1,...,L]$. The pixels are separated into two classes $C_0$ for the background and $C_1$ for objects by a threshold level~$t$. Pixels with intensity levels $[1,...,t]$ are classified into $C_0$ where each pixel is assigned a value 0, while pixels with intensity levels $[t+1,...,L]$ belong to $C_1$ where each pixel is assigned value 1. Given an image $f$ with dimensions $height\times weight \times channels$, the pixel intensity of the $c^{th}$ channel at position $(x,y)$ is dentoted as $f_c(x,y)$, where $c\in \{R,G,B\}$. The adapted Otsu's thresholding is applied independently to each channel to calculate the optimal threshold values $t^*_R$, $t^*_G$, and $t^*_B$. For each channel $c$, the optimal threshold $t^*_c$ is determined by maximizing the between-class variance:

\begin{equation}
    \sigma^2_B{(t_c)} =  \frac{[\mu_T\omega(t_c)-\mu(t_c)]^2}{\omega(t_c)[1-\omega(t_c)]}.
\end{equation}
Above, $t_c$ varies over all possible intensity values for channel~$c$, $\sigma^2_B{(t_c)}$ represents the between-class variance for a given threshold $t$ in channel $c$. Subsequently, the optimal threshold $t^*_c$ is: 

\begin{equation}
    \sigma^2_B{(t^*_c)} = \max_{1 \leq t_c < L} \sigma^2_B{(t_c)}.
\end{equation}

Each channel $c$ of the image $f(x,y)$ is then segmented into $C_1$ and $C_0$ according to its respective threshold value $t^*_c$:

\begin{equation}
    g_c(x,y) = \begin{cases}
        1 & if f_c(x,y) > t^*_c \\
        0 & \textit{otherwise.}
    \end{cases}
\end{equation}

Here, $g_c(x,y)$ denotes the Boolean segmented output for channel $c$ at position $(x,y)$. The Boolean outputs for the RGB channels are combined to reconstruct the segmented image. This approach enables the application of Otsu's thresholding to color images by considering the unique characteristics of each color channel.

\subsection{Color Thermometers}

Color thermometers use thermometer encodings, which is a form of input discretization that converts continuous input into a Boolean representation that preserves order \cite{buckman_thermometer_2018}. An image consists of $height \times width \times channels$, with each pixel intensity ranging within the discrete interval $[0,...,255]$. The color thermometer encoding technique begins with establishing a parameter $t$, which defines the number of thresholds to be applied across the intensity spectrum of each color channel. Each threshold is calculated based on an equal division of the~$[0,...,255]$ intensity range, creating $t$ evenly spaced values that serve as cut-off points for Boolean encoding. An example of pixels being mapped using thermometer encoding is shown in Table \ref{tab:thermometer-encoding}.

\bgroup
\def\arraystretch{1.05}
\begin{table}[ht]
\centering
\caption{Example of Pixel Intensity Values Mapped to Thermometer Codes Using Eight Evenly Spaced Levels \cite{buckman_thermometer_2018}.}
\label{tab:thermometer-encoding}
\begin{tabular}{l l} 
 \hline
Pixel Value & Discretized (thermometer) \\
 \hline
 15 & [ 0 1 1 1 1 1 1 1 ]\\ 
 149 & [ 0 0 0 1 1 1 1 1 ]\\ 
 255 & [ 0 0 0 0 0 0 0 0 ]\\ 
 \hline
\end{tabular}
\end{table}
\egroup

For each color channel, the method generates $t$ Boolean values. A Boolean value of 0 indicates that the pixel's intensity exceeds the corresponding threshold, while a 1 denotes the opposite. This process transforms each original RGB pixel into a $3\times t$-dimensional Boolean vector, where each subset of $t$ bits represents the color thermometer encoding for one of the RGB channels.

\subsection{Adaptive Color Thermometers}

The multilevel thresholding algorithm segments an image into multiple levels \cite{arora_multilevel_2008}. This algorithm was combined with the existing thermometer encoding algorithm \cite{buckman_thermometer_2018}, which uses evenly distributed threshold values ranging from $[0,...,255]$. The reason for combining these two algorithms was that it was assumed that placing evenly distributed thresholds would sometimes misplace the threshold values on peaks in the histogram of the image. The multilevel thresholding algorithm starts by calculating the mean and standard deviation of all pixels in the range $R=[a,...,b]$. Where the initial values are $a=0$ and $b=255$. Then calculate the sub-range boundaries $T_1$ and $T_2$:
\begin{equation}
    T_1 = \mu - k \cdot \sigma,
    \label{t_1}
\end{equation}
\begin{equation}
    T_2 = \mu + k \cdot \sigma,
    \label{t_2}
\end{equation}
where $\mu$ and $\sigma$ are the mean and standard deviation of the pixels in the range $R$ and $k$ is a free parameter. Pixels possessing intensity values within the intervals $[a,...,T_1]$ and $[T_2,...,b]$ are assigned threshold values that correspond to the weighted averages of their respective intensity values. Then $a=T_1+1$ and $b=T_2-1$. The aforementioned steps are repeated $\frac{n}{2}-1$ times, where $n$ is equal to the amount of thresholds. Finally, the boundaries of the sub-range, as defined by \eqref{t_1} and \eqref{t_2}, are updated such that $T_1=\mu$ and $T_2=\mu + 1$. Subsequently, pixels with intensity values within the intervals $[a,...,T_1]$ and $[T_2,...,b]$ are assigned threshold values. These threshold values are determined by calculating the weighted averages of the intensity values within their respective intervals. This process segments the image into distinct levels using $n$ thresholds. These thresholds are then used for thermometer encoding.

\section{Empirical Results}
\label{sec:results}
In this section, we evaluate the performance of individual TM Specialists and TM Composites on the publicly available image classification dataset CIFAR-10. In the experiments, we change the clause size and scale the feedback threshold~($T$) thereafter. Here, we focus on the accuracy achieved and performance. The accuracy of the TM Specialists and TM Composites is measured by taking the mean of the last 25 epochs with sample variance. All experiments are run on an NVIDIA DGX H100 GPU. The source code for this paper is available at the Centre for Artificial Research's GitHub\footnote{https://github.com/cair/An-Optimized-Toolbox-for-Advanced-Image-Processing-with-Tsetlin-Machine-Composites}.

\subsection{Augmented and Non Augmented Dataset}
 The CIFAR-10 dataset consists of $60~000$ color images, evenly distributed across ten distinct classes: bird, cat, deer, dog, frog, horse, ship, truck, airplane, and automobile, each class containing $6~000$ images \cite{krizhevsky_learning_2012}. Furthermore, CIFAR-10 is split into $50~000$ training images and $10~000$ test images. In both datasets, a form of image augmentation was introduced to potentially reduce overfitting. Each image in the original training set was flipped horizontally, increasing the number of images to~$100~000$.
We use both the non-augmented and augmented version for training the TM Specialists. The TM Specialists trained on the augmented dataset comprising~$100~000$ images are denoted with the prefix "Augmented". 

\subsection{Hyperparameter Search}

In Table \ref{tab:hyperparameters-tm}, the hyperparameters used to train the TM Specialists for the first experiment are listed. The hyperparameters were found using Optuna \cite{akiba_optuna_2019} and through manual experimentation. Optuna was configured with the sampling algorithm Tree-structured Parzen Estimator Sampler (TPESampler) for this hyperparameter search. Here, each TM with the different image processing techniques was run five epochs, for 50 trials, where the combination of hyperparameters yielding the highest accuracy was selected for training. The hyperparameters in question are specificity ($s$), feedback threshold ($T$), convolution window, and whether the clauses should be weighted or not, found individually for each image processing technique utilized. For simplicity, the hyperparameter search was configured with the hyperparameters delineated in Table~\ref{tab:optuna}. 

\bgroup
\def\arraystretch{1.05}
\begin{table}[ht]
\centering
\caption{Hyperparameter Values Used for Hyperparameter Search for the Tsetlin Machine Specialists With Optuna.}
\label{tab:optuna}
\begin{tabular}{l l} 
\hline
Hyperparameter & Values \\
 \hline
 Convolution Window & $[1,32]$\\ 
 Weighted Clauses & $\{True, False\}$\\ 
 Clauses & $2~000$\\
 T & $[1,3~000]$\\
 s & $[1,10]$\\
 \hline
\end{tabular}
\end{table}
\egroup

Notice that we only used $2~000$ clauses in the hyperparameter search as evident in Table \ref{tab:hyperparameters-tm}, while we in Table \ref{tab:accuracy} employ from $2~000$ to $64~000$ clauses. As seen, we reuse the hyperparameters found for $2~000$ clauses for the remaining configurations. Also, we used the same hyperparameters for the non-augmented and augmented data. 

\begin{table*}
    \centering
    \caption{Tsetlin Machine Specialist Hyperparameters of the First Experiment After Hyperparameter Search.}
    \begin{tabular}{ l c c c c c c }
    \hline
    \noalign{\vskip 0.3mm} 
    TM Specialist & Epochs & Clauses & Convolution Window & Weighted Clauses & T & s\\
    \hline
        10x10 Canny Edge Detection & 250 & 2000 & 10x10 &  True & 1500 & 10.0 \\
        10x10 Augmented Canny Edge Detection & 250 & 2000  & 10x10 & True & 1500 & 10.0 \\
        5x5 Adaptive Gaussian Thresholding & 250 & 2000  & 5x5 & True & 2500 & 3.5 \\
        5x5 Augmented Adaptive Gaussian Thresholding & 250 & 2000  & 5x5 & True & 2500 & 3.5 \\
        10x10 Adaptive Mean Thresholding & 250 & 2000  & 10x10 & True & 500 & 10.0 \\
        10x10 Augmented Adaptive Mean Thresholding & 250 & 2000  & 10x10 & True & 500 & 10.0 \\
        10x10 Otsu's Thresholding & 250 & 2000  & 10x10 & True & 3000 & 10.0 \\
        10x10 Augmented Otsu's Thresholding & 250 & 2000  & 10x10 & True & 3000 & 10.0 \\
        3x3 Color Thermometers & 250 & 2000  & 3x3 & True & 3000 & 5.0 \\
        4x4 Color Thermometers & 250 & 2000  & 4x4 & True & 3000 & 5.0 \\
        5x5 Color Thermometers & 250 & 2000  & 5x5 & True & 3000 & 5.0 \\
        3x3 Augmented Color Thermometers & 250 & 2000  & 3x3 & True & 3000 & 5.0 \\
        4x4 Augmented Color Thermometers & 250 & 2000  & 4x4 & True & 3000 & 5.0 \\
        5x5 Augmented Color Thermometers & 250 & 2000  & 5x5 & True & 3000 & 5.0 \\
        Histogram of Oriented Gradients & 250 & 2000 & 32x32 & False & 50 & 10.0 \\
        Augmented Histogram of Oriented Gradients & 250 & 2000  & 32x32 & False & 50 & 10.0 \\
        3x3 Adaptive Color Thermometers & 250 & 2000  & 3x3 & True & 3000 & 5.0 \\
        4x4 Adaptive Color Thermometers & 250 & 2000  & 4x4 & True & 3000 & 5.0 \\
        5x5 Adaptive Color Thermometers & 250 & 2000  & 5x5 & True & 3000 & 5.0 \\
        3x3 Augmented Adaptive Color Thermometers & 250 & 2000  & 3x3 & True & 3000 & 5.0 \\
        4x4 Augmented Adaptive Color Thermometers & 250 & 2000  & 4x4 & True & 3000 & 5.0 \\
        5x5 Augmented Adaptive Color Thermometers & 250 & 2000  & 5x5 & True & 3000 & 5.0 \\
    \hline
    \end{tabular}
    \label{tab:hyperparameters-tm}
\end{table*}

 From the hyperparameter search, we can see that the most important hyperparameter values for TM accuracy are the size of the convolution window, $T$, and $s$. This is consistent with the findings in \cite{tarasyuk_systematic_2023}, which stated that $T$ and $s$ are the most important hyperparameters for influencing the accuracy of a TM. However, their research was carried out on the MNIST dataset, which is made up of black and white handwritten numbers.  CIFAR-10 consists of color images of animate and inanimate objects, which is a more complex task to classify for a TM. In addition, the convolution window is important in regards to which image processing technique is utilized. As such, there are differences in the size of the convolution window between the different TM Specialists depending on which image processing technique is utilized.  

\begin{table*}
\centering
\caption{Tsetlin Machine Specialists and Tsetlin Machine Composites Accuracy ($\Bar{x} \pm s^2$) on CIFAR-10 for Increasing Clause Sizes. Mean and Variance are Calculated for the 25 Last Epochs.}
\label{tab:accuracy}
\begin{tabular}{ l c c c c c c } 
\hline
\noalign{\vskip 0.26mm} 
TM Specialist & 2,000& 4,000& 8,000& 16,000& 32,000& 64,000\\
 \hline
 10x10 Canny Edge Detection & $48.7 \pm 0.22$ & $50.4 \pm 0.13$ & $53.4 \pm 0.11$ & $56.6 \pm 0.06$ & $58.8 \pm 0.06$ & $60.6 \pm 0.03$ \\
 10x10 Augmented Canny Edge Detection & $50.6 \pm 0.20$ & $51.5 \pm 0.21$ & $53.2 \pm 0.29$ & $56.5 \pm 0.21$ & $59.9 \pm 0.09$ & $62.4 \pm 0.06$ \\
 5x5 Adaptive Gaussian Thresholding & $57.9 \pm 0.16$ & $60.6 \pm 0.17$ & $63.0 \pm 0.11$ & $66.7 \pm 0.10$ & $68.2 \pm 0.03$ & $70.2 \pm 0.03$ \\ 
 5x5 Augmented Adaptive Gaussian Thresholding & $57.8 \pm 0.24$ & $60.3 \pm 0.15$ & $63.3 \pm 0.11$ & $65.9 \pm 0.04$ & $69.0 \pm 0.04$ & $70.3 \pm 0.02$ \\ 
 10x10 Adaptive Mean Thresholding & $57.8 \pm 0.14$ & $60.3 \pm 0.12$ & $62.3 \pm 0.05$ & $65.4 \pm 0.08$ & $67.2 \pm 0.06$ & $68.4 \pm 
0.13$ \\
 10x10 Augmented Adaptive Mean Thresholding & $59.1 \pm 0.20$ & $61.4 \pm 0.16$ & $63.4 \pm 0.18$ & $65.7 \pm 0.06$ & $68.3 \pm 0.06$ & $70.2 \pm 0.06$ \\
 10x10 Otsu's Thresholding & $56.9 \pm 0.13$ & $58.5 \pm 0.19$ & $61.2 \pm 0.17$ & $62.9 \pm 0.12$ & $63.9 \pm 0.23$ & $65.0 \pm 0.20$ \\
 10x10 Augmented Otsu's Thresholding & $58.5 \pm 0.17$ & $60.1 \pm 0.21$ & $61.5 \pm 0.23$ & $63.2 \pm 0.20$ & $65.2 \pm 0.19$ & $65.7 \pm 0.31$ \\
 3x3 Color Thermometers & $63.6 \pm 0.15$ & $65.8 \pm 0.42$ & $68.4 \pm 0.35$ & $70.4 \pm 0.26$ & $72.3 \pm 0.22$ & $74.3 \pm 0.15$ \\ 
 4x4 Color Thermometers & $64.6 \pm 0.16$ & $66.6 \pm 0.33$ & $68.9 \pm 0.44$ & $71.5 \pm 0.30$ & $73.6 \pm 0.31$ & $75.1 \pm 0.32$ \\
 5x5 Color Thermometers & $64.5 \pm 0.20$ & $66.8 \pm 0.17$ & $\mathbf{69.1 \pm 0.63}$ & $\mathbf{71.7 \pm 0.27}$ & $\mathbf{73.7 \pm 0.22}$ & $\mathbf{75.4 \pm 0.09}$ \\
 3x3 Augmented Color Thermometers & $64.1 \pm 0.19$ & $65.4 \pm 0.52$ & $67.5 \pm 0.86$ & $69.1 \pm 1.26$ & $70.5 \pm 1.86$ & $72.7 \pm 0.96$ \\
 4x4 Augmented Color Thermometers & $\mathbf{65.4 \pm 0.22}$ & $66.4 \pm 0.46$ & $68.5 \pm 0.52$ & $69.7 \pm 1.49$ & $71.6 \pm 1.07$ & $73.6 \pm 1.44$ \\
 5x5 Augmented Color Thermometers & $65.1 \pm 0.48$ & $\mathbf{66.9 \pm 0.46}$ & $68.6 \pm 0.81$ & $69.8 \pm 1.87$ & $71.7 \pm 2.15$ & $74.1 \pm 1.35$ \\
 Histogram of Oriented Gradients & $64.2 \pm 0.06$ & $65.7 \pm 0.04$ & $66.9 \pm 0.04$ & $67.3 \pm 0.03$ & $67.5 \pm 0.02$ & $67.5 \pm 0.02$ \\
 Augmented Histogram of Oriented Gradients & $64.8 \pm 0.07$ & $66.6 \pm 0.08$ & $67.5 \pm 0.03$ & $67.9 \pm 0.02$ & $68.4 \pm 0.03$ & $68.5 \pm 0.02$ \\
 3x3 Adaptive Color Thermometers & $60.8 \pm 0.10$ & $63.2 \pm 0.30$ & $65.3 \pm 0.14$ & $68.1 \pm 0.05$ & $70.3 \pm 0.06$ & $71.7 \pm 0.06$ \\
 4x4 Adaptive Color Thermometers & $61.3 \pm 0.30$ & $64.2 \pm 0.22$ & $66.9 \pm 0.25$ & $69.6 \pm 0.07$ & $71.5 \pm 0.07$ & $73.0 \pm 0.05$ \\
 5x5 Adaptive Color Thermometers & $62.0 \pm 0.25$ & $65.1 \pm 0.16$ & $67.2 \pm 0.14$ & $69.3 \pm 0.06$ & $71.6 \pm 0.06$ & $73.3 \pm 0.04$ \\
 3x3 Augmented Adaptive Color Thermometers & $61.3 \pm 0.25$ & $62.6 \pm 0.22$ & $64.8 \pm 0.26$ & $67.1 \pm 0.53$ & $68.9 \pm 0.45$ & $70.8 \pm 0.65$ \\
 4x4 Augmented Adaptive Color Thermometers & $62.6 \pm 0.23$ & $64.1 \pm 0.36$ & $66.2 \pm 0.21$ & $68.4 \pm 0.33$ & $70.3 \pm 0.75$ & $72.9 \pm 0.33$ \\
 5x5 Augmented Adaptive Color Thermometers & $62.9 \pm 0.21$ & $64.6 \pm 0.34$ & $66.4 \pm 0.39$ & $68.5 \pm 0.65$ & $71.1 \pm 0.40$ & $72.9 \pm 0.38$ \\
 Added Together in a TM Composite & $\mathbf{79.5 \pm 0.04}$ & $\mathbf{80.6 \pm 0.02}$ & $\mathbf{81.5 \pm 0.02}$ & $\mathbf{82.2 \pm 0.02}$ & $\mathbf{82.7 \pm 0.02}$ & $\mathbf{82.8 \pm 0.01}$ \\
\hline
\end{tabular}
\end{table*}

The highest accuracies achieved for each TM configuration are marked in bold in Table \ref{tab:accuracy}. The top performing TM Specialist is the 5x5 Color Thermometer trained with $64~000$ clauses with its accuracy of $75.4\%$. This accuracy is higher than the current state-of-the-art, which is 75.1\% \cite{sharma_drop_2023, granmo_tmcomposites_2023}. Furthermore, we discovered better hyperparameters for TMs using color thermometers and adaptive Gaussian thresholding algorithms than previously found in \cite{granmo_tmcomposites_2023}.

For each TM Specialist, we discovered robust and effective hyperparameters, indicated by accuracies that increase with clause pool size. This is in agreement with \cite{tarasyuk_systematic_2023}, which reports that when optimal $T$ and $s$ are found for a given number of clauses, adding more clauses yields higher accuracy. This can be seen for all of the TM Specialist except for Histogram of Oriented Gradients and Augmented Histogram of Oriented Gradients where the accuracy flattens after $32~000$ clauses.

Notably, the TM Composite comprised by the TM Specialists trained using $64~000$ clauses reaches an accuracy of~$82.8\%$. This TM Composite sets a new benchmark on CIFAR-10 for TMs, outperforming the previous state-of-the-art by $7.7$ percentage points. Also, notice the trend indicating that the accuracy of the TM Composite increases with the number of clauses employed by a TM Specialist up to $32~000$ clauses, with minimal increase after. This can be seen in Table \ref{tab:accuracy}, where the TM Composite consisting of TMs with~$32~000$ clauses achieves an accuracy of $82.7\%$ while the TM Composite with $64~000$ clauses has an accuracy of $82.8\%$. 

\begin{figure}[ht]
    \centering
        \includegraphics[width=\linewidth]{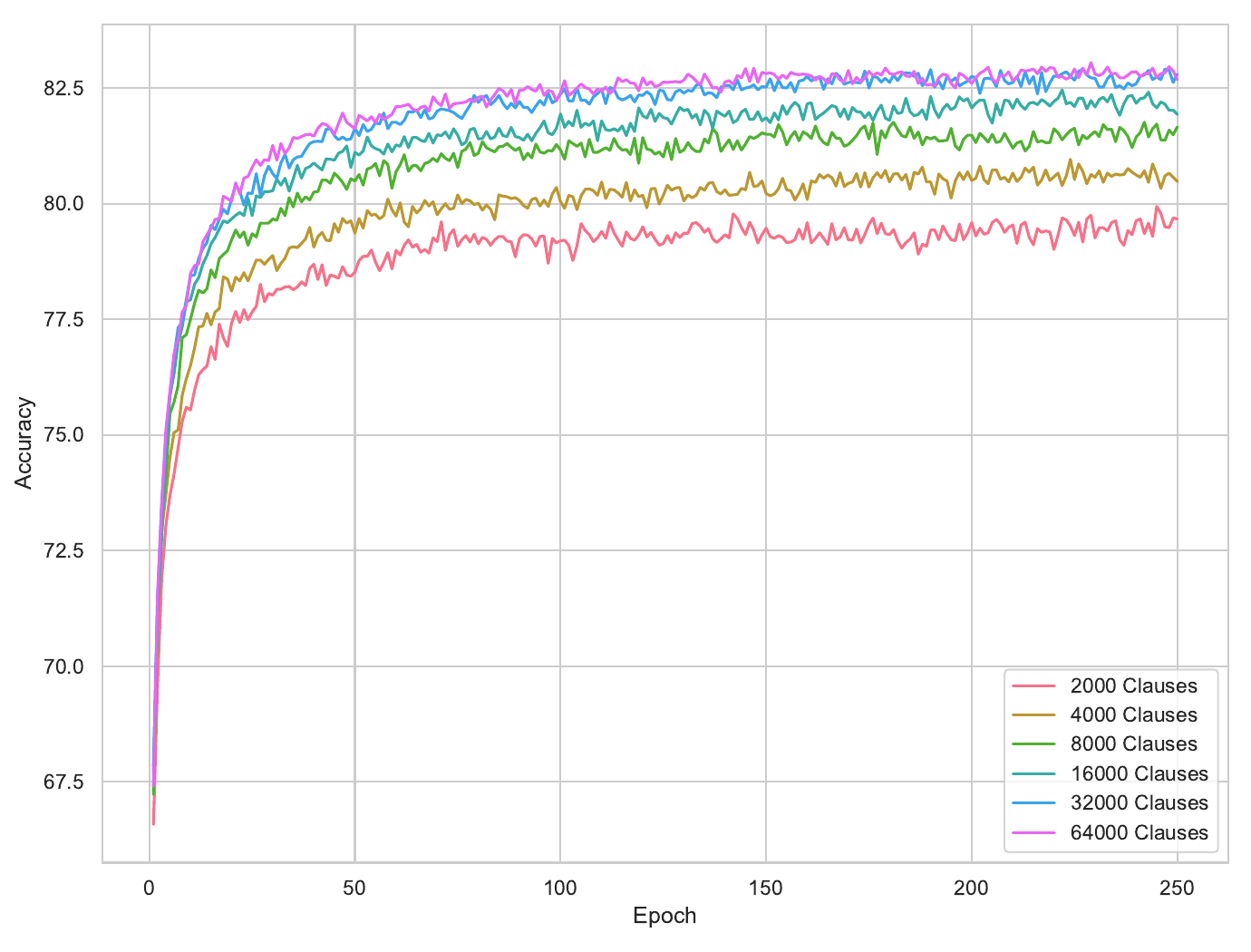}
        \caption{Tsetlin Machine Composites accuracy on the CIFAR-10 dataset, plotted epoch-by-epoch.}
    \label{fig:accuracy}
\end{figure}

In Fig.~\ref{fig:accuracy}, we can see that there is a faster increase in accuracy the more clauses the TM Specialists comprising the TM Composite have.

\section{Related Work}
\label{sec:related-work}
Concerning classification of color images using TMs, the recent key advances can be summarized as follows. Firstly, Mathisen and Smørvik \cite{mathisen_analysis_2020} performed an analysis of image processing techniques that use several different TM architectures for classification. They implemented Otsu's thresholding, adaptive Gaussian thresholding, and Canny edge detection for the original TM, Weighted TM, Convolutional TM (CTM), and Layered TM. To benchmark the different image processing techniques for the TM architectures, they utilized a modified CIFAR-10 dataset. In their analysis, the CTM achieved the highest accuracy of 60.7\% on CIFAR-10. In our paper, we also used Otsu's thresholding, adaptive Gaussian thresholding, and Canny edge detection for image processing. However, we utilized several more image processing techniques and the novel TM Composites architecture. Furthermore, we achieved higher accuracy on CIFAR-10 compared to their work. 
Sharma~et~al.~\cite{sharma_drop_2023} proposed a technique called Drop Clause for the TM that drops clauses randomly to avoid overfitting when learning patterns. Drop Clause reduces the training time as the TM has fewer clauses to check if the pattern matches. In addition, the Drop Clause technique for the TM using~$60~000$ clauses achieved an accuracy of 75.1\% on CIFAR-10. To Booleanize the images they used the adaptive Gaussian thresholding technique and the TM architecture utilized is the CTM. In our paper, we also utilize the adaptive Gaussian thresholding technique and the CTM. However, we implement several more image processing techniques and utilize the novel TM Composites architecture. Additionally, we achieved greater accuracy on CIFAR-10. Granmo \cite{granmo_tmcomposites_2023} proposed the TM Composites architecture using $2~000$ clauses per TM Specialist, which achieved an accuracy of 75.1\% on CIFAR-10. The TM Composites architecture is used in this paper and is explained in more detail in Section \ref{subsec:tmcomposite}. 
The four TM Specialists utilized HOG, adaptive Gaussian thresholding, and color thermometers. 
However, we utilize more image processing techniques and conduct a hyperparameter search for the TM Specialists. Finally, we achieve greater accuracy on CIFAR-10 with an accuracy of 82.8\%.

\section{Conclusions and Further Work}
\label{sec:conclusion}
In this paper, we present an image preprocessing approach and several image processing techniques for TM Specialists that enhance the accuracy when all are added together in a TM Composite. Furthermore, we conduct a rigorous hyperparameter search using Optuna that significantly improves the accuracy of several TM Specialists. The result is a toolbox that provides new state-of-the-art results for TMs on CIFAR-10 with an accuracy of 82.8\%. For future work, we are interested in looking into whether the aforementioned approach will achieve better results on datasets other than CIFAR-10. In addition, there are several image processing techniques not tested for TM Specialists that could be interesting to explore.

\bibliographystyle{IEEEtran}
\bibliography{references}

\end{document}